\PassOptionsToPackage{table}{xcolor} % Pass table option to xcolor

\documentclass[wcp]{jmlr} % W&CP article
\usepackage[utf8]{inputenc}

\usepackage{silence}
\usepackage[font=small,labelfont=bf]{caption}

\usepackage{graphicx}
\usepackage{xcolor} % now loaded only once, with the table option passed above
\usepackage{booktabs}
\usepackage{multirow}
\usepackage{float}    % For [H] option
\usepackage{enumitem}
\usepackage{subcaption}
\usepackage[T1]{fontenc}

\definecolor{bronze}{rgb}{0.8, 0.5, 0.2}

\usepackage{booktabs}
\usepackage{graphicx}
\usepackage[table]{xcolor}  % for color in tables
\usepackage{booktabs}
\usepackage{multirow}
\usepackage{caption}  % For caption centering
\usepackage{float}    % For [H] option
\usepackage{tcolorbox}
\usepackage{caption}
\theorembodyfont{\upshape}
\theoremheaderfont{\scshape}
\theorempostheader{:}
\theoremsep{\newline}

\jmlrproceedings{AABI 2025}{Proceedings of the 7th Symposium on Advances in Approximate Bayesian Inference, 2025}

\title{Post-Hoc Uncertainty Quantification in Pre-Trained Neural Networks via Activation-Level Gaussian Processes}
 \author{\Name{Richard Bergna} \Email{rsb63@cam.ac.uk}\\
 \addr Department of Engineering, University of Cambridge
 \AND
 \Name{Stefan Depeweg}\\
 \addr Siemens AG\\
\Name{Sergio Calvo-Ordoñez}\\
  \addr Mathematical Institute, University of Oxford \\
  \Name{Jonathan Plenk}\\
  \addr Mathematical Institute, University of Oxford \\
  \Name{Alvaro Cartea}\\
  \addr Mathematical Institute, University of Oxford\\
  \Name{Jose Miguel Hernández-Lobato}\\
  \addr Department of Engineering, University of Cambridge
 }

\begin{document}

\maketitle

\begin{abstract}
Uncertainty quantification in neural networks through methods such as Dropout, Bayesian neural networks and Laplace approximations is either prone to underfitting or computationally demanding, rendering these approaches impractical for large-scale datasets. In this work, we address these shortcomings by shifting the focus from uncertainty in the weight space to uncertainty at the activation level, via Gaussian processes. More specifically, we introduce the Gaussian Process Activation function (GAPA) to capture neuron-level uncertainties. Our approach operates in a post-hoc manner, preserving the original mean predictions of the pre-trained neural network and thereby avoiding the underfitting issues commonly encountered in previous methods. We propose two methods. The first, GAPA-Free, employs empirical kernel learning from the training data for the hyperparameters and is highly efficient during training. The second, GAPA-Variational, learns the hyperparameters via gradient descent on the kernels, thus affording greater flexibility.  Empirical results demonstrate that GAPA-Variational outperforms the Laplace approximation on most datasets in at least one of the uncertainty quantification metrics.
\end{abstract}

% Keywords may be removed
%\begin{keywords}
%List of keywords
%\end{keywords}

\section{Introduction}
\begin{figure}[htbp]
    \centering
    \includegraphics[width=0.45\linewidth]{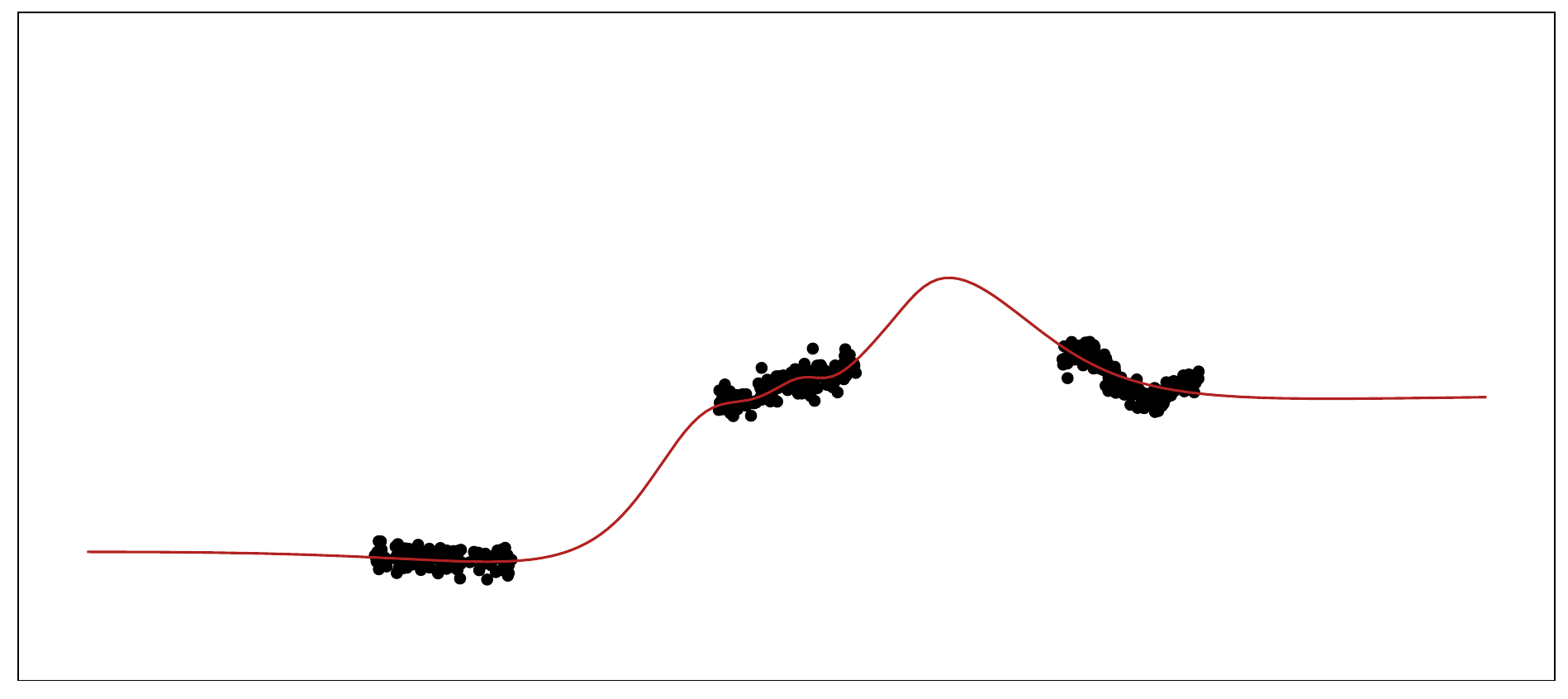}
    \includegraphics[width=0.45\linewidth]{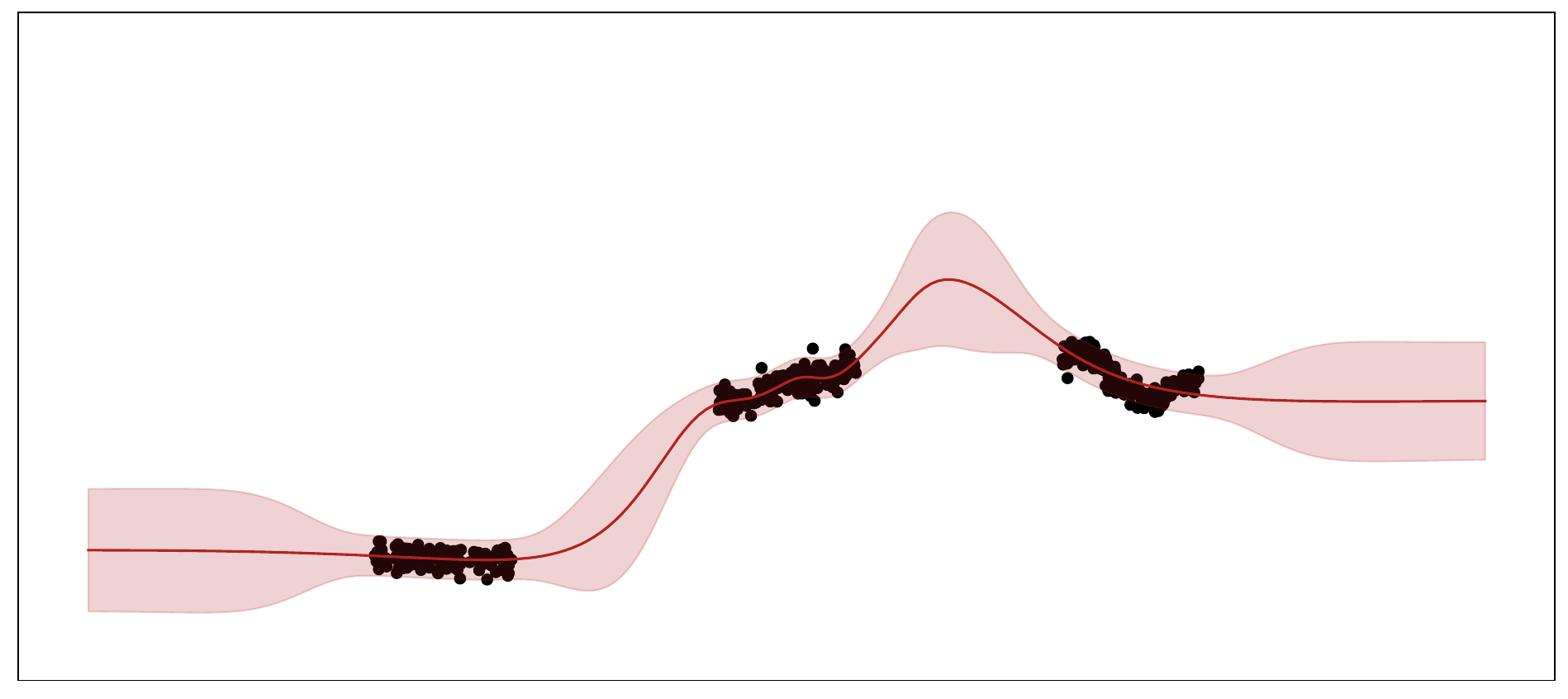}
    \caption{(Left) The architecture of the pre-trained backbone neural network. (Right) The GAPA module, applied post-hoc to the first layer to quantify uncertainty without modifying the original predictions. Illustration based on a toy regression problem from \citep{ortega2023variational}.}

    \label{fig:example}
\end{figure}

Deep neural networks (DNNs) have achieved state-of-the-art performance in a wide range of pattern recognition tasks \citep{krizhevsky2012imagenet, kenton2019bert, mnih2015human, hinton2012deep, litjens2017survey}. However, traditional DNNs do not quantify epistemic uncertainty, limiting their reliability in risk-sensitive applications such as autonomous driving \citep{shafaei2018uncertainty}, healthcare \citep{begoli2019need}, and finance \citep{blasco2024survey}. To address this limitation, numerous surrogate methods have been developed for downstream decision-making under uncertainty, particularly for anomaly detection and out-of-distribution detection \citep{li2023rethinking, liu2023gen}. Yet, a more principled Bayesian approach has been proposed to model uncertainty directly. This has led to methods that approximate distributions over weight space, including Bayesian Neural Networks \citep{neal2012bayesian}, deep ensembles \citep{lakshminarayanan2017simple}, and Markov Chain Monte Carlo methods. Additionally, regularization-based methods such as Dropout \citep{gal2016dropout} and SWAG \citep{maddox2019simple}, as well as explicit modeling of weight uncertainty \citep{blundell2015weight}, have shown promise in improving uncertainty estimates in deep learning models.
However, there are many challenges that hinder the widespread applications of Bayesian modelling: In general, these methods are computationally expensive or even intractable in practice, for instance requiring the training of multiple DNNs or learning a distribution over each weight \citep{graves2011practical,hernandez2015probabilistic}. With the rise of large pre-trained models in many domains like computer vision and natural language, the need to incorporate uncertainty-aware methods already during the model training phase is another limiting factor in their applications \citep{fort2019deep,izmailov2021bayesian}. Even methods, such as Monte-Carlo dropout, which may be present during training to act as a regularizer, require multiple forward passes to generate samples \citep{gal2016dropout, neal2012bayesian, lakshminarayanan2017simple}. In addition, many Bayesian methods tend to suffer from underfitting, because uncertainty modelling is often inherently linked to regularization (mostly via the prior) \citep{wenzel2020good,osawa2019practical}.
Recently, Laplace approximations have become popular, arguably because they can be applied as a post-processing method to a pre-trained neural network without affecting its prediction and empirically capture uncertainty well without requiring sampling. Nevertheless, they demand the calculation of the Jacobian, which is computationally intensive. In addition, for scalability reasons, they are typically only employed in the last layer of a model, which potentially hinders their flexibility \citep{daxberger2021laplace, ortega2023variational}.

In this work, we approach this problem from a different perspective: \textbf{What if we shift our focus from uncertainty in the weight space to uncertainty in the activations?} Specifically, we model uncertainty at each neuron's postactivation by fitting a one-dimensional Gaussian process to each neuron in the first layer.  This approach is inexpensive to fit, and can be applied to pre-trained neural networks, without the need of re-training or fine-tuning.  The second key ingredient is to propagate the obtained a uncetainties at the GP-infused layer (GAPA) through the network using deterministic propagation rules akin to determistic variational inference \citep{wu2018deterministic}. Unlike Laplace approximation this combination allows us to model uncertainty at any layer of the network.  The method is purely post-hoc (it only needs access to the pre-trained model and some training data), does not require fine-tuning of the model, and, unlike for instance dropout, can express uncertainty in a single forward-pass. Importantly, infusing uncertainty in this way does not change the original prediction of the pre-trained model in any way, thereby preserving the models predictive quality.

Specifically, we propose two Gaussian Process Activation function (GAPA) methods. The first, GAPA-Free, is a cost-effective approach that employs empirical kernel methods to compute the hyperparameters of the Gaussian process. The second, GAPA-Variational, uses variational inducing points to learn the hyperparameters, thereby allowing for greater flexibility. Our contributions are as follows:

\begin{itemize}[label=\textbullet, leftmargin=*] 
\item A post-hoc method for pre-trained neural networks that extends them through uncertainty modelling without affecting their predictions. 
\item A delta approximation method to propagate the uncertainty from the activation space to the output space. 
\item Empirical demonstration that GAPA—and in particular GAPA-Variational—delivers exceptional performance in uncertainty quantification, outperforming Laplace approximations on most datasets.
\item A novel approach to uncertainty quantification by focussing on modelling the uncertainty at the activation level rather than in the weight space. 
\end{itemize}

% \begin{itemize}

%     \item modelling uncertainty is often in effect both regularization and uncertainty modelling. For example, techniques like dropout, ensembling or VI modify predictions by incorporating uncertainty. This  often leads to under-fitting

%      \item In classical UQ models, pre-trained models cannot be used. Ensembling, SWAG, Dropout, VI, MCMC all require that they are used during training

%     \item Ideally, we seek a method that performs only uncertainty modelling without regularization. We leave the regularization efforts to the specifics of the training. E.g. via drop-out (non Monte Carlo), weight decay etc. This means the uncertainty-modelling is disentangled from regularization.

%     \item We also seek a method that can be applied post-hoc. That means taking a pre-trained model and extending it through uncertainty modelling without affecting its prediction. 

%     \item This method should be scalable and easy to use. Its training should also be cheaper than the original training of the pre-trained models. E.g. not requiring gradient descent.

%     \item Given these limitations, existing methods for uncertainty modelling are inadequate for our requirements.
    
%     \item These limitations motivate our proposed method.

%     \item The combination of learning Gaussian processes on the activations (post-hoc) and DVI to propagate the uncertainty, while keeping the mean/original propagation fixed fullfills all these properties.

%     \item GAPA can be seen as  keeping a finger print of the traning data in the activation levels,...
% \end{itemize}

\section{Model Proposition: GAPA + Uncertainty Propagation}

We begin by presenting the GAPA method, which aims to quantifiy uncertainty in a pre-trained neural network. We assume the network was first trained in a supervised manner on a dataset \(\mathcal{D} = \{ (\mathbf{x}_n, \mathbf{y}_n) \}_{n=1}^N\). Then, to estimate uncertainty, we augment the network by applying a Gaussian Process (GP) to the output of each neuron in  a layer. To highlight the generality of the approach we assume here, that this method is applied to the first hidden layer of the network.
Figure~\ref{fig:example} illustrates the backbone network and the GAPA module.

\subsection{Pretrained Neural Network}

Consider a standard feedforward neural network with \(L\) layers:  For \(l=0,\dots,L\), the \((l+1)\)-th layer contains \(D_l\) neurons with weight matrix \(W^l \in \mathbb{R}^{D_l \times D_{l-1}}\), biases $b^l \in \mathbb{R}^{D_l}$ and activation function \(a^l\). For an input \(x\in\mathbb{R}^{D_0}\), the network's prediction is given by
\[
\hat{y}_{\mathbf{x}} = W^L a^L\Bigl(W^{L-1}a^{L-1}\bigl(\cdots a^1(W^0x + b^0) \cdots \bigr) + b^{L-1}\Bigr) + b^L.
\]
This pre-trained network is optimised using standard supervised learning on \(\mathcal{D}\), and its parameters are subsequently fixed.

\subsection{GAPA: Gausisan Process Activations}

\begin{figure}[htbp]
    \centering
    \includegraphics[width=0.32\linewidth]{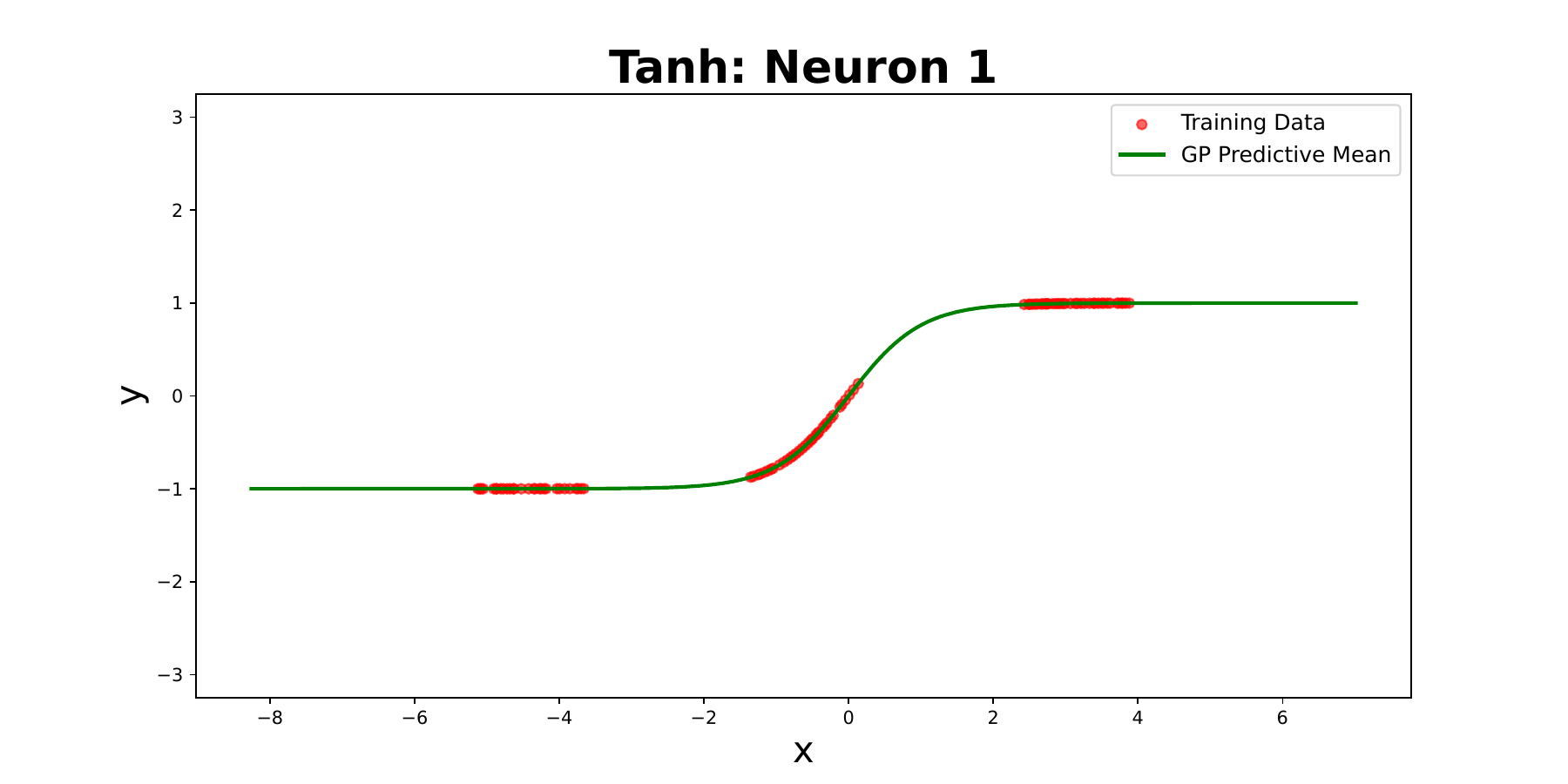}
    \includegraphics[width=0.32\linewidth]{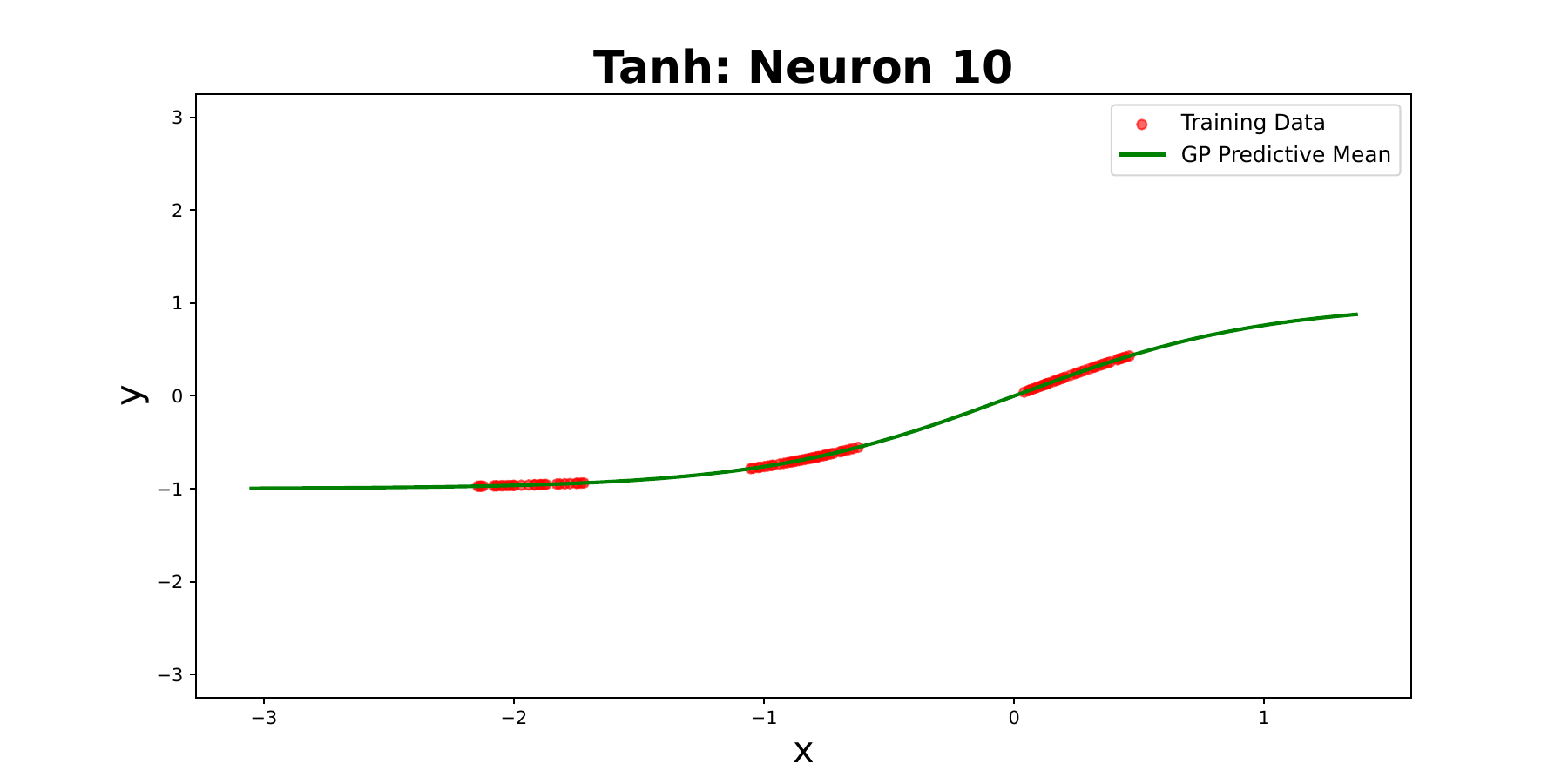}
    \includegraphics[width=0.32\linewidth]{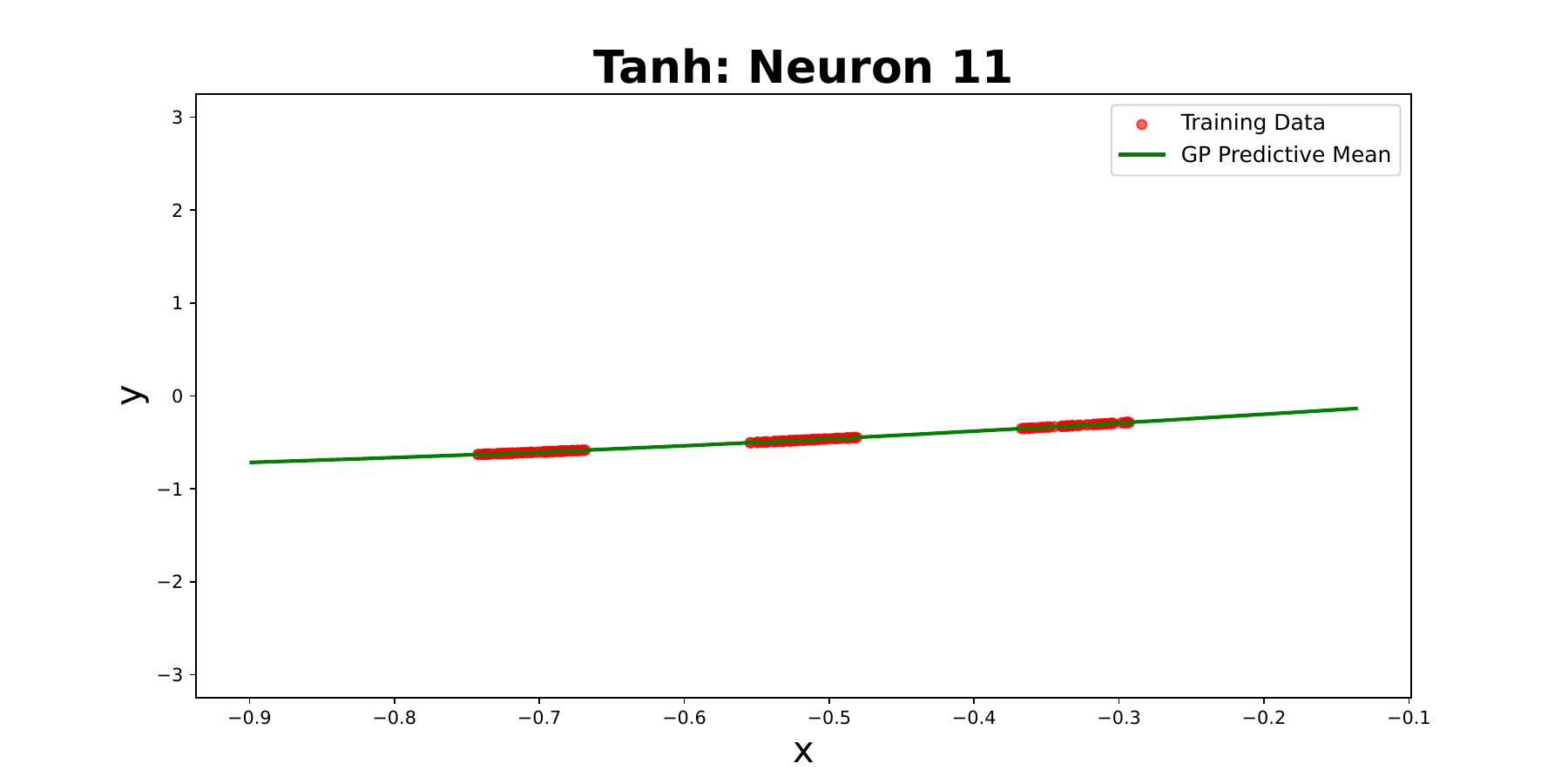}
    \includegraphics[width=0.32\linewidth]{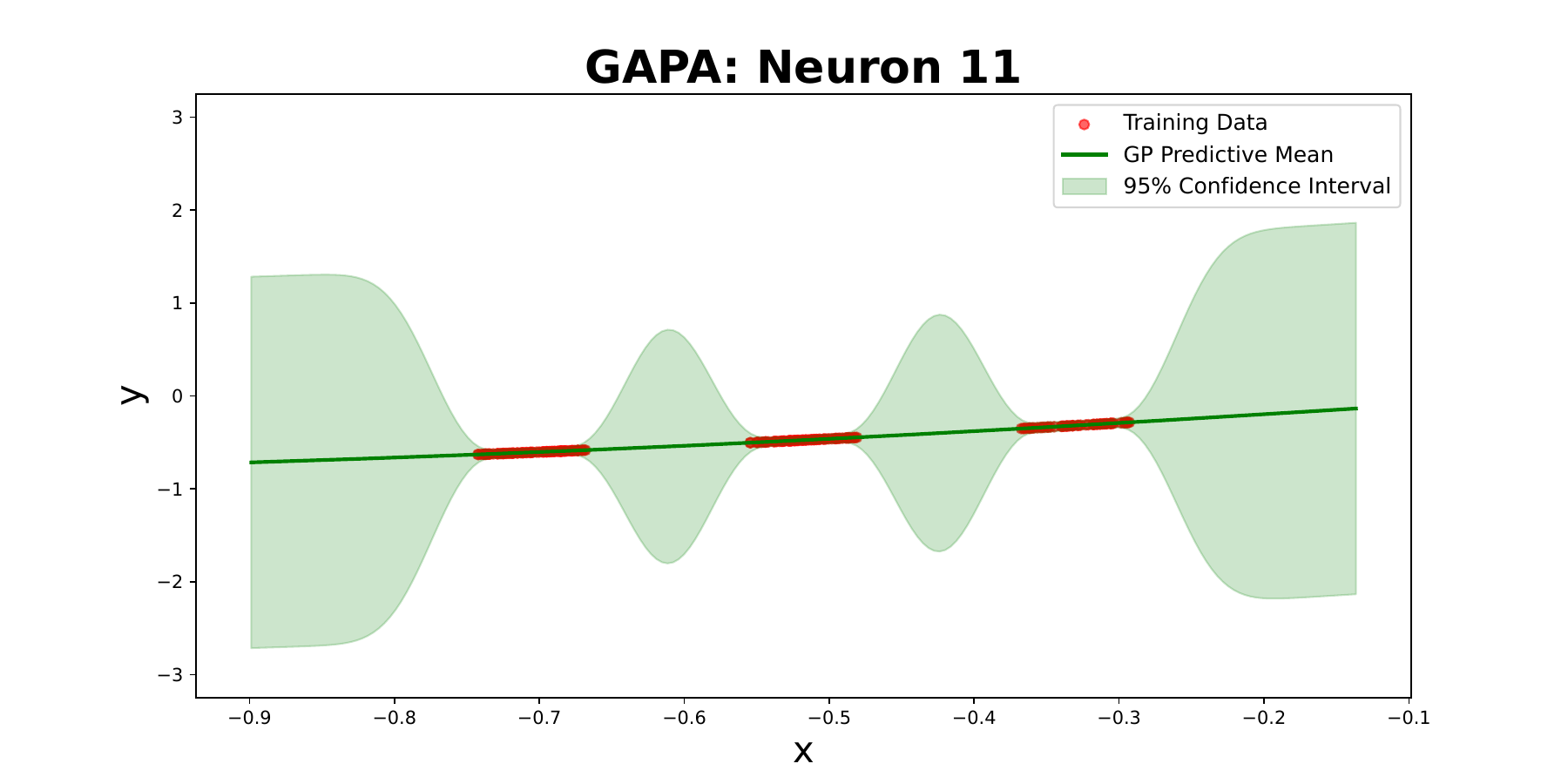}
    \includegraphics[width=0.32\linewidth]{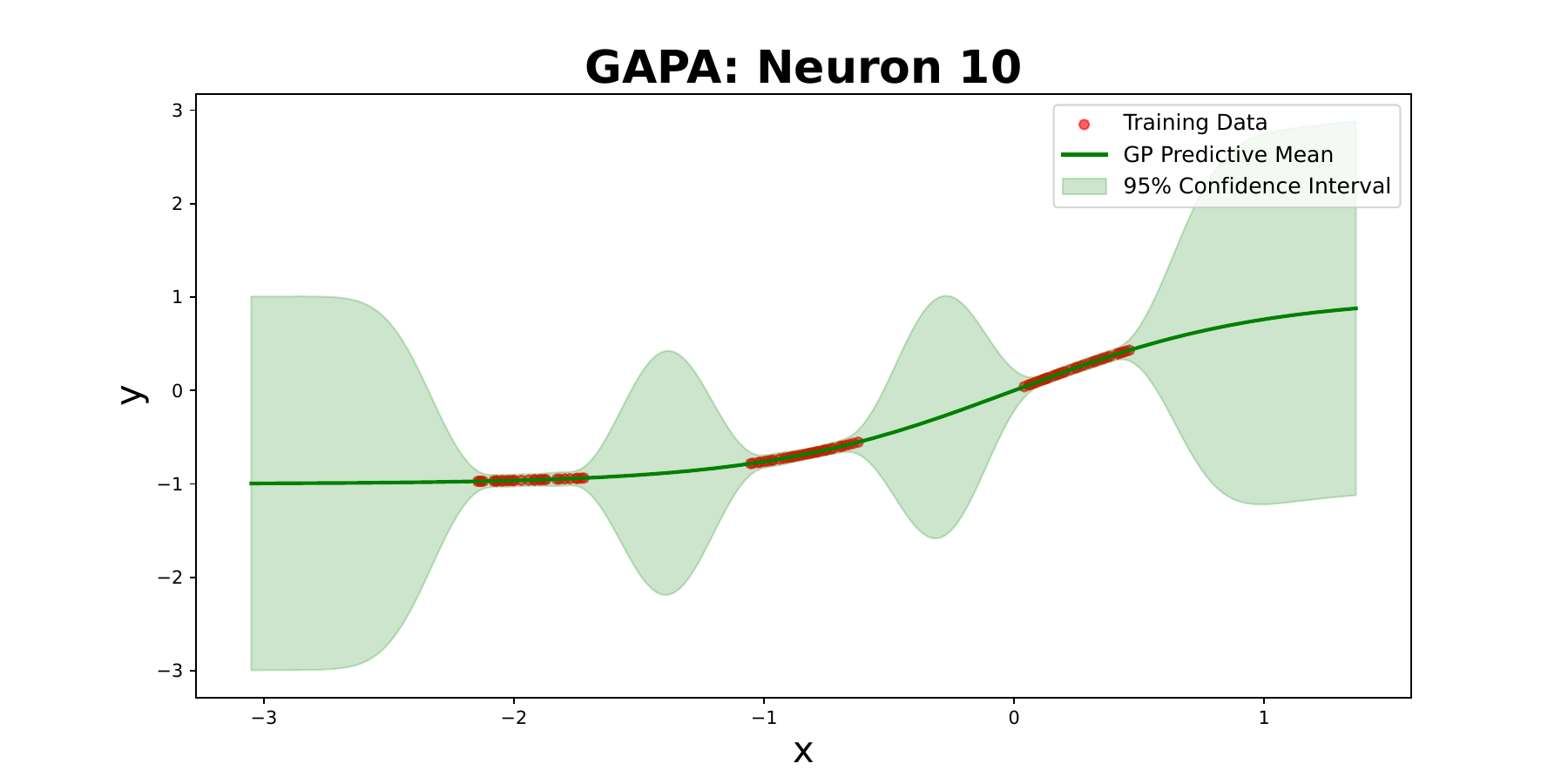}
    \includegraphics[width=0.32\linewidth]{images/gapa_neuron_11.pdf}
    \caption{Baseline activations (Top) versus GAPA activations (bottom) for neurons 1, 10, 11. GAPA preserves the mean activation while providing an uncertainty estimate.}
    \label{fig:activations}
\end{figure}

To quantify the uncertainty of a pre-trained network without affecting its mean predictions, we attach an independent one-dimensional GP to each neuron in the first layer. Here, the pre-trained network (with fixed parameters) has been optimised on the dataset \(\mathcal{D} = \{ (\mathbf{x}_n, \mathbf{y}_n) \}_{n=1}^N\) using standard supervised learning. Let \(X := W^0x + b^0 \in \mathbb{R}^{D^1}\) denote the neurons of the first layer. For \(d \in \{1,\dots,D^1\}\), let \(Y_d := a^1(X_d)\) be the activation of the \(d\)-th neuron. We introduce uncertainty at the activation-level by replacing $a^1(X_d)$ with a GP $f_d(X_d) + \epsilon_d$. Here, we assume a GP prior \(f_d \sim \mathcal{GP}(m_d, k_d)\), with mean function 
$
m_d(X_d) := a^1(X_d),
$
and a covariance kernel \(k_d\) (specifically, the RBF kernel with hyperparameters learned via an empirical method; see Appendix~\ref{Appendix:empirical_kernel} for further details).
Denote the neurons and activations of the training data at the first layer by \(\mathbf{X}\) and \(\mathbf{Y}_d = a^1(\mathbf{X}_d)\).
The posterior mean is computed as
\[
\mu_d(X_d) = m_d(X_d) + k_d(X_d, \mathbf{X}_d)\Bigl[K_d(\mathbf{X}_d,\mathbf{X}_d) + \sigma_n^2 I_N\Bigr]^{-1}\Bigl(\mathbf{Y}_d - m_d(\mathbf{X}_d)\Bigr).
\]
As we have \(Y_d = m_d(X_d)\) by construction, it follows that $\mu_d(X_d) = m_d(X_d) = a^1(X_d).$

Hence the pre-trained network's original activation is preserved. The posterior covariance
\[
\Sigma_d(X_d,X'_d) = k_d(X_d,X'_d) - k_d(X_d, \mathbf{X}_d)\Bigl[K_d(\mathbf{X}_d,\mathbf{X}_d) + \sigma_n^2 I_N\Bigr]^{-1} k_d(\mathbf{X}_d,X'_d),
\]
quantifies the epistemic uncertainty in the $d$-th neuron's activation. Note, that this doesn't depend on the prior mean. As shown in Figure~\ref{fig:activations} for neurons 1, 10, and 11, the GAPA model preserves the baseline activations while adding a principled uncertainty estimate. In summary, by using a GP whose prior mean is set equal to the neuron's true activation (i.e. its label), we preserve the pre-trained network’s mean predictions while simultaneously providing a rigorous uncertainty epistemic estimate via the GP’s posterior covariance.

% Don't think we should open the aleatoric uncertainty box, here
% Further, by conditioning on $f_d = a^1$, the GP activation models aleatoric uncertainty through
% \begin{equation}
%     (f_d(X_d) + \epsilon_d) | (f_d = a^1) \sim \mathcal{N}(a^1(X_d), \sigma^2).
% \end{equation}

\subsection{Propagating the Variance through the Network}
Since the GP at the first layer is constructed to preserve the pre-trained network’s mean activations, the mean forward pass remains identical to that of the pre-trained model. We now need to define a variance-forward path. For this we identify two scenarios: linear layers (such as in dense and convolutional layers) and non-linear activation functions. 

\paragraph{Linear Transformation of Variance.} Since a linear transformation of a Gaussian remains Gaussian, if the input variance is \(\Sigma_a\) and the linear layer applies a transformation \(z = W a\), then the resulting variance is given by $\Sigma_z = W\,\Sigma_a\,W^\top.$

\paragraph{Propagation Rules for Non-Linear Activations.}
For a non-linear activation \(y = g(z)\) applied to a Gaussian random variable \(z \sim \mathcal{N}(\mu, \sigma^2)\), we approximate \(g(z)\) by a first-order Taylor expansion (delta approximation) 
\[
g(z) \approx g(\mu) + g'(\mu)(z-\mu).
\]
Since \(z-\mu \sim \mathcal{N}(0, \sigma^2)\), this yields an approximate variance of
$
\operatorname{Var}(y) \approx \left(g'(\mu)\right)^2 \sigma^2.
$

% For instance, if \(g(z)=\tanh(z)\), then \(g'(\mu)=1-\tanh^2(\mu)\) and the propagated variance becomes
% \[
% \operatorname{Var}(y) \approx \left(1-\tanh^2(\mu)\right)^2 \sigma^2.
% \]

% While the delta approximation is universally applicable, there also exist alternativ approximation techniques, e.g. for \texttt{ReLU} activations, we can use a moment-matching approach under a rectified Gaussian distribution.

% \textbf{TODO.} Say that one assumes $g(z)$ is Gaussian, for the approximations in further layers.

\paragraph{Overall Variance Propagation.} By sequentially applying the linear transformation rule for variance and the delta approximation for non-linear activations, we obtain a tractable, layer-wise method for propagating uncertainty from the first layer (where the GP is applied) to the final network output.

\subsection{GAPA-Free: Linear Scaling of the Output Variance}
After propagating uncertainty to the network output, we refine the variance using a simple linear transformation:
\[
\operatorname{Var}_{\text{final}} = \theta_1\, \operatorname{Var}_{\text{output}} + \theta_2,
\]
where \(\theta_1\) (a scaling factor) and \(\theta_2\) (an offset) are learned to capture any residual uncertainty. This calibration is computationally efficient since it involves only two parameters and requires no additional backpropagation through the network.

%via gradient descent. As only two parameters are involved, this linear scaling is very computationally efficient.

\subsection{GAPA-Variational}
In GAPA-Variational, rather than applying a fixed linear scaling, the GP variational parameters (similar to those used in variational GPs \citep{titsias2009variational}) are optimized via maximum likelihood. For each neuron \(d\), we assume a GP prior $f_d \sim \mathcal{GP}\bigl(m_d, k_d\bigr)$
with \(m_d(X_d)=a^1(X_d)\) (i.e. the neuron's activation) and a covariance kernel \(k_d\) (e.g. the RBF kernel with empirically determined hyperparameters). We introduce inducing variables \(\mathbf{u}_d\) with fixed inducing inputs \(\mathbf{Z}_d=\mathbf{X}_d\) (taken from the training data of the first layer) and set the inducing mean to \(m_d(\mathbf{Z}_d)=a^1(\mathbf{Z}_d)\). The corresponding variational distribution is defined as
$
q(\mathbf{u}_d) = \mathcal{N}\bigl(m_d(\mathbf{Z}_d),\, S_d\bigr),
$
where \(S_d\) (the variational covariance) and the kernel hyperparameters \(\theta_d\) are learned. Let \(y_i\) denote the target for the \(i\)th input, and let \(\mu_i\) and \(\sigma_i^2\) be the predictive mean and variance obtained by propagating the GP uncertainties through the network (using, for example, the delta approximation). Because the GP prior mean is fixed to the pre-trained activation, the posterior mean remains unchanged and only the uncertainty (variance) is learned. Consequently, the overall training objective is the Gaussian negative log-likelihood (NLL)
$
    \mathcal{L} = \sum_{i=1}^N \frac{1}{2}\log\bigl(2\pi\sigma_i^2\bigr) + \frac{(y_i - \mu_i)^2}{2\sigma_i^2}.
$

This loss function is optimized by backpropagating the NLL from the network's final output while keeping the pre-trained network weights fixed. In this way, GAPA-Variational provides a flexible, data-driven uncertainty estimate through the learned GP covariance, all while preserving the original mean predictions of the pre-trained network.

% \subsection{Delta approximation from activation to output uncertainty}

% Suppose we wish to propagate uncertainty through a non-linear activation function, for example, the hyperbolic tangent (\(\tanh\)) or the rectified linear unit (ReLU). Let \(z\) be a Gaussian random variable with mean \(\mu\) and variance \(\sigma^2\), i.e.,
% \[
% z \sim \mathcal{N}(\mu, \sigma^2).
% \]
% Consider an activation function \(g(\cdot)\) (such as \(\tanh\) or ReLU), and let the transformed variable be
% \[
% y = g(z).
% \]

% Using the \emph{delta method}, we approximate the propagation of uncertainty by linearising \(g(z)\) around the mean \(\mu\). A first-order Taylor expansion of \(g(z)\) about \(\mu\) gives
% \[
% g(z) \approx g(\mu) + g'(\mu)(z - \mu),
% \]
% where \(g'(\mu)\) is the derivative of \(g(\cdot)\) evaluated at \(\mu\).

% Since \(z - \mu\) is a centred Gaussian with variance \(\sigma^2\), the linearised variable \(y\) is approximately Gaussian with mean
% \[
% \mathbb{E}[y] \approx g(\mu)
% \]
% and variance
% \[
% \operatorname{Var}[y] \approx \left(g'(\mu)\right)^2
% \]
% In practice, a smoothed or approximate derivative can be used for propagation of uncertainty.

% Thus, the delta variational inference approach provides a computationally efficient method to approximate how uncertainty in the Gaussian (representing the neuron's pre-activation uncertainty) propagates through a non-linear activation function, yielding an approximate Gaussian distribution for the neuron's output. This approximation is especially useful in scenarios where exact propagation is intractable.

\section{Results}

\begin{table}[htbp]
    \centering
    \caption{Results on regression datasets. 
    Best values are in \textcolor{purple}{purple}, 
    second-best in \textcolor{teal}{teal}, 
    and third-best in \textcolor{bronze}{bronze}.
    An asterisk (*) indicates a last-layer LLA variant.}
    \label{tab:results}
    \vspace{0.5em}  % Space between caption and table
    \resizebox{\textwidth}{!}{%
    \begin{tabular}{l|ccc|ccc|ccc}
    \toprule
    \multirow{2}{*}{\textbf{Model}} 
      & \multicolumn{3}{c|}{\textbf{Airline}} 
      & \multicolumn{3}{c|}{\textbf{Year}} 
      & \multicolumn{3}{c}{\textbf{Taxi}} \\
    \cmidrule(lr){2-4} \cmidrule(lr){5-7} \cmidrule(l){8-10}
    & \textbf{NLL} & \textbf{CRPS} & \textbf{CQM}
    & \textbf{NLL} & \textbf{CRPS} & \textbf{CQM}
    & \textbf{NLL} & \textbf{CRPS} & \textbf{CQM} \\
    \midrule
    MAP 
      & 5.087 
      & 18.436 
      & 0.158 
      & 3.674 
      & 5.056 
      & 0.164 
      & 3.763 
      & \textcolor{teal}{3.753} 
      & \textcolor{bronze}{0.227} \\
    LLA Diag 
      & 5.096 
      & \textcolor{teal}{18.317} 
      & 0.144 
      & 3.650 
      & 4.957 
      & 0.122 
      & 3.714 
      & 3.979 
      & 0.270 \\
    LLA KFAC 
      & 5.097 
      & \textcolor{teal}{18.317} 
      & 0.144 
      & 3.650 
      & 4.955
      & 0.121 
      & 3.705 
      & 3.977 
      & 0.270 \\
    LLA* 
      & 5.097 
      & \textcolor{bronze}{18.319} 
      & 0.144 
      & 3.650 
      & 4.954
      & 0.120 
      & 3.718 
      & 3.965 
      & 0.270 \\
    LLA* KFAC 
      & 5.097 
      & \textcolor{teal}{18.317} 
      & 0.144 
      & 3.650 
      & 4.954
      & 0.120 
      & 3.705 
      & 3.977 
      & 0.270 \\
    ELLA 
      & 5.086 
      & 18.437 
      & 0.158 
      & 3.674 
      & 5.056 
      & 0.164 
      & 3.753 
      & \textcolor{bronze}{3.754} 
      & \textcolor{bronze}{0.227} \\
    VaLLA 100 
      & \textcolor{teal}{4.923} 
      & 18.610 
      & \textcolor{teal}{0.109} 
      & \textcolor{teal}{3.527} 
      & 5.071 
      & 0.084
      & \textcolor{bronze}{3.287} 
      & 3.968 
      & \textcolor{teal}{0.188} \\
    VaLLA 200 
      & \textcolor{purple}{4.918} 
      & 18.615 
      & \textcolor{purple}{0.107} 
      & \textcolor{purple}{3.493} 
      & 5.026 
      & \textcolor{teal}{0.076}%  <-- Example for 3rd best 
      & \textcolor{teal}{3.280} 
      & 3.993 
      & 0.188 \\
    \hline
    \textbf{GAPA-Free}
      & 5.083
      & 18.394
      & 0.115
      & 3.644
      & \textcolor{teal}{4.909}% <-- Example for 3rd best CRPS
      & \textcolor{bronze}{0.084}
      & 3.668
      & 4.01
      & 0.274 \\
    \textbf{GAPA-Variational}
      & \textcolor{bronze}{5.067}
      & \textcolor{purple}{18.282}
      & \textcolor{bronze}{0.135}
      & \textcolor{bronze}{3.545}% <-- Example for 3rd best NLL
      & \textcolor{purple}{4.796}% <-- Example for best CRPS
      & \textcolor{purple}{0.053}% <-- Example for best CQM
      & \textcolor{purple}{3.268}
      & \textcolor{purple}{3.552}
      & \textcolor{purple}{0.154} \\
    \bottomrule
    \end{tabular}
    }
    \vspace{0.5em}
\end{table}
We compare GAPA’s predictive distribution with state-of-the-art Laplace-based methods for post-hoc uncertainty quantification in pre-trained networks—including VaLLA, LLA variants, and ELLA \citep{daxberger2021laplace, izmailov2020subspace, ortega2023variational}—on three benchmark regression datasets: (i) the UCI Year dataset, (ii) the US flight delay (Airline) dataset \citep{dutordoir2020flight}, and (iii) the Taxi dataset \citep{salimbeni2017deep}. We follow the original train/test splits used in prior studies.

Table~\ref{tab:results} summarizes the performance of our proposed models compared to state-of-the-art post-processing methods on several regression datasets. Our evaluation metrics include Negative Log-Likelihood (NLL), Continuous Ranked Probability Score (CRPS) \citep{gneiting2007strictly}, and the Centered Quantile Metric (CQM) \citep{ortega2023variational}. In the table, the best values are highlighted in \textcolor{purple}{purple}, the second-best in \textcolor{teal}{teal}, and the third-best in \textcolor{bronze}{bronze}.
Our experimental results show that both GAPA-Free and GAPA-Variational achieve competitive performance. Notably, GAPA-Variational consistently enhances uncertainty quantification. For example, on the \emph{Airline} dataset, it attains the best CRPS while its NLL and CQM values rank among the top three. On the \emph{Year} dataset, GAPA-Variational records the best CRPS and CQM scores with a competitive NLL. Most importantly, on the \emph{Taxi} dataset, it outperforms all other methods across all metrics. These findings indicate that our approach successfully propagates uncertainty from the activation space to the network's final output without altering the pre-trained network's predictions. As a result, GAPA-Variational preserves the base network's predictive accuracy while providing a more reliable and nuanced uncertainty estimate, making it well suited for risk-sensitive applications.

% These results indicate that our approach successfully propagates uncertainty from the first layer through to the final output without altering the pre-trained network's mean predictions. The GAPA-Variational model not only retains the predictive accuracy of the base network but also delivers a more reliable and nuanced estimation of uncertainty, making it a compelling choice for risk-sensitive applications.

\section{Related Work}

In \cite{morales2020activation}, auNN replaces activations with GPs and trains them jointly across layers via variational inference, requiring multiple samples at inference time. In contrast, our method uses the original activation (e.g., ReLU) as the GP prior mean---thereby preserving the pre-trained network's predictions---and fits GPs solely to quantify the uncertainty of the activation function. This post-hoc approach avoids re-training the network and achieves uncertainty estimation with a single forward pass.

\section{Conclusion}
In this work, we have introduced the Gaussian Process Activation function (GAPA), a novel framework designed to quantify uncertainty in pre-trained neural networks. We have also presented a theoretically principled method to propagate uncertainty from the activations space to the output space using the delta approximation approach. Our approach empirically outperforms the Laplace approximation method, achieving faster training times. Nevertheless, Gaussian processes remain computationally expensive at inference time. Future work will focus on exploring scalable models or approximations to Gaussian processes to optimise computational efficiency, as well as extending the model to classification task.

\bibliography{references}

\appendix

\section{Empirical Estimation of Inducing Inputs and RBF Kernel Hyperparameters}\label{Appendix:empirical_kernel}

\paragraph{Inducing Input Selection:} To set the RBF kernel hyperparameters in a data-driven manner, we first select inducing inputs for each neuron's GP based on the empirical cumulative distribution function (CDF) of its pre-activation values. Let \(x\) denote the one-dimensional pre-activation values for a given neuron, and assume these values are sorted as
\[
x_{(1)} \leq x_{(2)} \leq \cdots \leq x_{(N)}.
\]
The empirical CDF is then given by
\[
F(x_{(i)}) = \frac{i}{N}, \quad i=1,\dots,N.
\]
To robustly capture the data distribution—especially the boundaries critical for out-of-distribution detection—we always include the minimum \(x_{(1)}\) and maximum \(x_{(N)}\) as inducing points. The remaining inducing inputs are selected by partitioning the CDF into equal quantile intervals. Specifically, if \(M\) inducing points are desired (with two reserved for the minimum and maximum), then the other \(M-2\) inducing points correspond to quantile levels
\[
p_m = \frac{m+1}{M-1}, \quad m = 1, 2, \dots, M-2.
\]
Each inducing input is chosen as the \(x_{(i)}\) whose empirical CDF value is closest to the corresponding \(p_m\).

\paragraph{RBF Kernel Hyperparameter Estimation:}  
The RBF kernel is defined as
\[
k(x, x') = \sigma_f^2 \exp\!\left(-\frac{(x - x')^2}{2\ell^2}\right),
\]
where:
\begin{itemize}
    \item \(\ell\) is the lengthscale, and
    \item \(\sigma_f^2\) is the output scale (variance constant).
\end{itemize}
We estimate the lengthscale \(\ell\) as a chosen quantile (e.g., the 25th percentile) of the pairwise Euclidean distances among the selected inducing inputs:
\[
\ell = \operatorname{quantile}\Bigl(\{|x_i - x_j| : i \neq j\},\, q\Bigr), \quad \text{with } q = 0.25.
\]
The output scale is set based on the variance of the training outputs (activation function):
\[
\sigma_f^2 = \max\Bigl(1,\, \operatorname{Var}(y_{\text{train}})\Bigr).
\]

\end{document}